\pdfoutput=1
\documentclass[10pt,twocolumn,letterpaper]{article}

\usepackage[pagenumbers]{cvpr}

\newcommand{\TODO}[1]{\textbf{\color{red}[T\if\relax\detokenize{#1}\relax\else: #1\fi]}}

\usepackage{xspace}
\newcommand{\ours}{TaCo\xspace}
\newcommand{\oursT}{TaCo-T\xspace}
\newcommand{\oursS}{TaCo-S\xspace}
\newcommand{\oursB}{TaCo-B\xspace}
\newcommand{\rv}{RV1126B\xspace}

\usepackage{booktabs}
\usepackage{multirow}
\usepackage{makecell}
\usepackage{siunitx}
\usepackage{subcaption}
\usepackage{colortbl}
\usepackage{pifont}
\definecolor{oursrow}{gray}{0.92}
\newcommand{\best}[1]{\textbf{#1}}
\definecolor{rankfirst}{RGB}{204,0,0}
\definecolor{ranksecond}{RGB}{0,82,204}
\definecolor{rankthird}{RGB}{38,153,38}
\newcommand{\first}[1]{\textcolor{rankfirst}{#1}}
\newcommand{\second}[1]{\textcolor{ranksecond}{#1}}
\newcommand{\third}[1]{\textcolor{rankthird}{#1}}
\newcommand{\na}{--}
\newcommand{\cmark}{\ding{51}}
\newcommand{\xmark}{\ding{55}}
\makeatletter
\renewcommand{\paragraph}[1]{\vspace{.25em}\noindent\textbf{#1\ifnum\spacefactor>\@m\else.\fi}}
\makeatother
\DeclareMathOperator{\RoIAlign}{RoIAlign}
\DeclareMathOperator{\BN}{BN}

\definecolor{cvprblue}{rgb}{0.21,0.49,0.74}
\usepackage[pagebackref,breaklinks,colorlinks,allcolors=cvprblue]{hyperref}
\usepackage{bibunits}
\makeatletter
\AtBeginDocument{%
  \let\orig@Hy@backout\Hy@backout
  \def\Hy@backout#1{{\let\@auxout\@mainaux\orig@Hy@backout{#1}}}%
  \let\orig@BR@starttoc\BR@starttoc
  \def\BR@starttoc{\@ifundefined{tf@brf@open}{\orig@BR@starttoc}{}}%
  \let\orig@BR@backref\BR@backref
  \def\BR@backref#1{\orig@BR@backref{#1\@extra@b@citeb}}%
  \def\endbibunit{%
    \global\let\cite\std@cite
    \global\let\@citex\std@@citex
    \global\let\@citew\std@@citew
    \global\let\nocite\std@nocite
    \@finishbibunit
    \gdef\@extra@b@citeb{}%
    \gdef\@extra@binfo{}%
    \@input{bu.aux}%
  }%
}
\newenvironment{suppbib}{%
  \gdef\@extra@b@citeb{@supp}%
  \gdef\@extra@binfo{@supp}%
  \bibunit[ieeenat_fullname]%
}{%
  \endbibunit
}
\newcommand{\corrfntext}[1]{%
  \begingroup
    \def\@thefnmark{*}%
    \@ifundefined{H@@footnotetext}{\@footnotetext{#1}}{\H@@footnotetext{#1}}%
  \endgroup
}
\makeatother

\def\confName{CVPR}
\def\confYear{2027}

\title{TaCo: Rethinking Correlation for Visual Tracking on Edge Devices}

\author{Qingmao Wei$^{1,2}$ \qquad Fagui Liu$^{1,2,*}$ \qquad Quan Tang$^{2,*}$\\
$^{1}$South China University of Technology \qquad $^{2}$Pengcheng Laboratory\vspace{-0.4em}
}

\begin{document}
\maketitle
\corrfntext{Corresponding authors: Quan Tang ({\ttfamily\scriptsize tangq@pcl.ac.cn}), Fagui Liu.}

\begin{abstract}
Lightweight trackers have replaced cross-correlation with attention, and on low-power edge hardware they pay for it, since memory-bound softmax, normalization and reshaping operators of attention cost far more time than their arithmetic suggests.
Benchmarking efficient trackers on an edge Neural Processing Unit (NPU) and an Arm CPU under one numerically validated protocol, we find that correlation trackers are the fast ones, but they lag in accuracy unless they adopt a ViT backbone, which brings attention back.
We revisit what correlation lacks.
Template--search interaction serves two roles: \emph{conditioning} the search features on the target and \emph{matching} them against it spatially.
Attention does both, whereas correlation only matches, and does so on features extracted without knowledge of the target.
We present \ours (\textbf{Ta}rget-aligned \textbf{Co}rrelation), which supplies the missing conditioning without attention by letting the template enter the per-frame network only as operands computed once per video.
A descriptor of the target scales and shifts the search features channel by channel inside the backbone, and kernels aligned with the target box, each with a weight predicted from the template, correlate with them in one matrix product.
With the backbone fixed, our correlation remains essential while incurring negligible NPU overhead, offering a better speed-accuracy trade-off than the attention-based interactions we tested.
\oursT achieves 62.3\% LaSOT AUC at 184\,FPS on the Rockchip \rv NPU and 96\,FPS on a Raspberry Pi 5 CPU, 3.0$\times$ and 3.5$\times$ the NPU speed of HiT-Tiny and AsymTrack-T at higher accuracy.
\oursB reaches 65.6\% at 50\,FPS on the NPU, more than twice the speed of HiT-Base, AsymTrack-B and LightFC-vit, again at higher accuracy.
Code and models are available at \mbox{\textnormal{\small\url{https://github.com/TsingWei/TaCo-Tracker}}}.
\end{abstract}

\section{Introduction}
\label{sec:intro}

\begin{figure}[t]
  \centering
  \includegraphics[width=\linewidth]{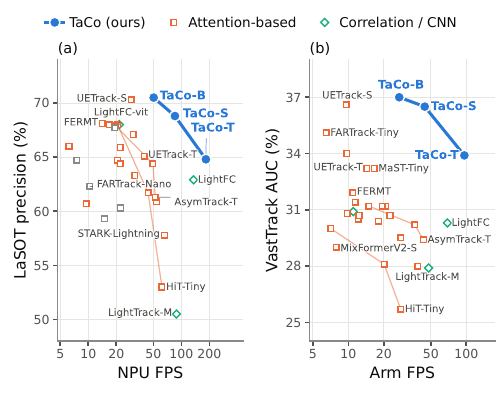}
  \caption{\textbf{\ours extends the accuracy and speed frontier on two edge devices.}
  (a) LaSOT precision vs.\ per-frame graph FPS on the Rockchip \rv NPU.
  (b) VastTrack AUC vs.\ per-frame graph FPS on the Raspberry Pi 5 Arm CPU.
  Both FPS axes are logarithmic. Lines connect variants within a tracker family, and gray markers fail the NPU FP16 numerical check.}
  \label{fig:teaser}
\end{figure}

Interaction between template and search features is central to visual tracking.
Siamese correlation trackers compare template and search features explicitly after backbone extraction, keeping inference lightweight on convolutional architectures~\cite{2016_siamfc,2019_siamrpnpp,2024_lightfc}, though they lag behind on distractors and appearance changes.
Attention-based trackers replaced explicit correlation with learned feature fusion~\cite{2021_transt,2021_stark}, and one-stream designs integrated interaction into feature extraction~\cite{2022_ostrack,2022_mixformer}.
Efficient trackers follow this route via compact transformers, distillation, and token pruning~\cite{2023_hit,2023_mixformerv2,2024_smat,2026_uetrack}, or cache the template to skip re-encoding~\cite{2024_litetrack,2025_asymtrack}.

\begin{figure}[t]
  \centering
  \includegraphics[width=\linewidth]{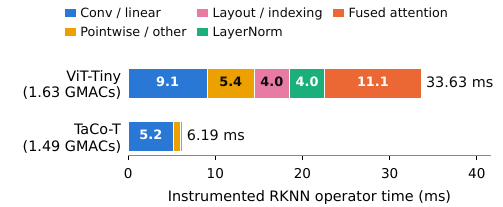}
  \caption{\textbf{Operator profile comparing ViT-Tiny and \oursT.}
  Matched warm RKNN operator profiles (\rv, FP16, 950\,MHz, librknnrt~2.3.2) comparing a ViT-Tiny tracker against our convolutional \oursT graph at a comparable MAC budget.}
  \label{fig:vit_taco_profile}
\end{figure}

Multiply-accumulate operations (MACs) give only a theoretical bound on execution speed.
Prior efficient trackers often report uncompiled PyTorch timings of desktop GPU or CPU latency.
Under compiled edge runtimes, however, low MACs do not guarantee low latency. The cached-template tracker AsymTrack-T~\cite{2025_asymtrack} requires fewer MACs than the CNN correlation tracker LightFC~\cite{2024_lightfc}, yet runs substantially slower on both the Arm CPU and the edge NPU.
Specifically, although caching removes template re-encoding, search-side softmax attention, normalization, and layout operators still execute every frame.
On-device operator profiling confirms this (\cref{fig:vit_taco_profile}). Compared with ViT-Tiny, convolutional backbones at a comparable MAC budget sharply reduce \rv NPU latency, because most of the runtime falls outside convolution and linear layers into fused attention, LayerNorm, layout ops (\texttt{Transpose}/\texttt{Split}/\texttt{Reshape}), and pointwise ops in the transformer.
Since compiled edge graphs favor dense convolutions and matrix products, edge trackers should avoid softmax attention, normalization, and heavy layout reshaping in both backbone and interaction.
Beyond latency, compiled FP16 graphs can also diverge numerically from FP32 references. We therefore evaluate \rv NPU (RKNN) and Raspberry Pi 5 Arm CPU (ncnn) graphs under one unified protocol with FP32 numerical checks alongside re-evaluated accuracy across multiple benchmarks.

This motivates a closer look at why correlation trackers lag behind attention trackers.
Interaction between the template and search features serves two roles, \emph{conditioning} search feature extraction on the target and supplying \emph{spatial matching} evidence for localization.
Siamese trackers extract search features without target knowledge and apply correlation only near the head~\cite{2016_siamfc,2019_siamrpnpp,2024_lightfc}, whereas rescuing accuracy with a ViT backbone~\cite{2024_lightfc} forfeits edge-friendly operators.
Attention performs both roles in every layer and couples them inside costly ops, yet the two roles need not share an operator.
We ask whether a correlation tracker can gain target-aware features while retaining inexpensive explicit matching on edge devices.

We answer this with \ours (\textbf{Ta}rget-aligned \textbf{Co}rrelation), which separates the two roles across edge-friendly operators.
Using a CNN backbone truncated at stride~16, \ours computes all template-derived operands once per video. A box-weighted descriptor predicts channel-wise scales and shifts before the stride-16 search blocks, while box-aligned $\RoIAlign$ target kernels, pooled context kernels, and template-predicted kernel weights separate part-level foreground matching from background context and calibrate correlation responses without per-frame overhead.
Because search features are already target-aware, we drop post-correlation fusion blocks and share compact reparameterizable RepConv layers~\cite{technique_ding2021repvgg} across the head, folding training branches into plain convolutions at deployment.
Per frame, the template enters only as operands of channel-wise multiply-adds and a single matrix product, without softmax, token normalization, or attention reshapes.

Instantiating \ours across three CNN variants (\oursT with LowFormer-E1, \oursS with RepViT-M1.1, and \oursB with RepViT-M1.5) extends the accuracy and speed frontier on both devices (\cref{fig:teaser}).
On \rv and Raspberry Pi 5, the \ours family consistently outperforms prior lightweight trackers across standard benchmarks while scaling smoothly from ultra-fast real-time inference to higher-capacity models.
For instance, \oursT reaches 184\,FPS on \rv (over $3\times$ faster than HiT-Tiny and AsymTrack-T with higher accuracy), while \oursB attains 65.6\% LaSOT AUC at 50\,NPU FPS ($2.3\times$ faster than LightFC-vit at higher accuracy) alongside strong gains across TrackingNet, GOT-10k, and VastTrack.
Controlled ablations confirm that both decoupled roles are essential, and transplanting this interaction onto representative transformer, CNN, and hybrid backbones (HiT, LightFC, and AsymTrack-T) likewise preserves or improves accuracy at substantially lower edge latency.

We make three contributions:
\begin{itemize}
  \item \textbf{Rethinking correlation.} A controlled study and transfer analysis across backbones separate target-conditioned feature extraction from explicit spatial matching, establishing their complementary gains and measured NPU/CPU operator costs.
  \item \textbf{\ours, an edge tracking family.} We realize this division using template operands cached once per video (target-guided modulation, target-aligned kernels with cached kernel weighting, and a shared reparameterizable head) across three CNN backbones.
  \item \textbf{A unified edge benchmark.} We pair compiled \rv NPU and Raspberry Pi 5 Arm CPU graph timings, verified against FP32 numerical references, with re-evaluated accuracy across four tracking benchmarks.
\end{itemize}

\section{Related Work}
\label{sec:related}

\paragraph{Correlation-based tracking}
Siamese trackers match template and search representations through explicit cross-correlation after feature extraction~\cite{2016_siamfc,2018_siamrpn,2019_siamrpnpp}.
Pixel-wise correlation treats template feature vectors as individual $1{\times}1$ kernels, producing a similarity map for each template location~\cite{yan2021alpharefine,liao2020pgnet,guo2021siamgat}.
Lightweight designs reduce the cost of this pipeline through architecture search~\cite{2021_lighttrack}, compact fusion modules~\cite{2022_fear}, or streamlined pixel-wise matching~\cite{2024_lightfc,2025_lightfc_x}.
LightFC combines pixel-wise matching with channel attention and convolutional fusion, and its ViT variant uses a stronger feature extractor while retaining the matching pipeline~\cite{2024_lightfc}.
Target-aware feature extraction has also been explored. InBN~\cite{guo2022inbn} injects target information into Siamese backbone stages, while SBT~\cite{xie2022sbt} interleaves cross-image interaction with feature extraction and removes the separate correlation step.
These studies establish the relevance of where template information enters the tracking network.

\paragraph{Attention-based and efficient tracking}
TransT~\cite{2021_transt} and STARK~\cite{2021_stark} use attention to fuse features from convolutional backbones, while one-stream trackers such as OSTrack~\cite{2022_ostrack}, MixFormer~\cite{2022_mixformer}, and SimTrack~\cite{2022_simtrack} unify  interaction between template and search with feature extraction.
To reduce the resulting quadratic cost, efficient variants explore hierarchical or separable-attention transformers~\cite{2023_hit,2024_smat,2024_fermt,2026_uetrack}, alongside distillation, layer-adaptive or sparse inference, and fast autoregressive designs~\cite{2023_mixformerv2,2025_sglatrack,2026_mast,2026_fartrack}.
Template caching further avoids repeated encoding through asymmetric attention~\cite{2022_mixformer} or asynchronous feature extraction~\cite{2024_litetrack}.
Yet even when template encoding is cached or theoretical MACs are lowered, search-side softmax attention, normalization, and heavy token reshapes still execute every frame.
As a result, latency reductions reported on desktop GPUs or uncompiled PyTorch do not directly transfer to compiled edge runtimes, where non-convolutional operators dominate execution time and can introduce FP16 numerical divergence.

\paragraph{Feature-wise modulation}
FiLM~\cite{perez2018film} conditions neural features on external context through per-channel affine transformations.
For video segmentation, OSMN~\cite{yang2018osmn} predicts object-specific channel scales from the first annotated frame, using a separate modulator to specialize the segmentation network to the object of interest.
In tracking, AutoMatch~\cite{2021_automatch} evaluates FiLM as a matching operator after backbone feature extraction and retains it in both classification and regression branches.
PromptVT~\cite{2024_promptvt} introduces target guidance through visual prompts.
AsymTrack~\cite{2025_asymtrack} caches template representations and injects them into the search branch through prototype-based modulation, retaining a later attention layer for interaction.
\ours uses feature-wise modulation within the search backbone and retains explicit correlation for spatial matching, evaluating this division under the operator costs of edge inference.

\section{TaCo: Target-aligned Correlation}
\label{sec:method}

\begin{figure*}[t]
  \centering
  \includegraphics[width=\linewidth]{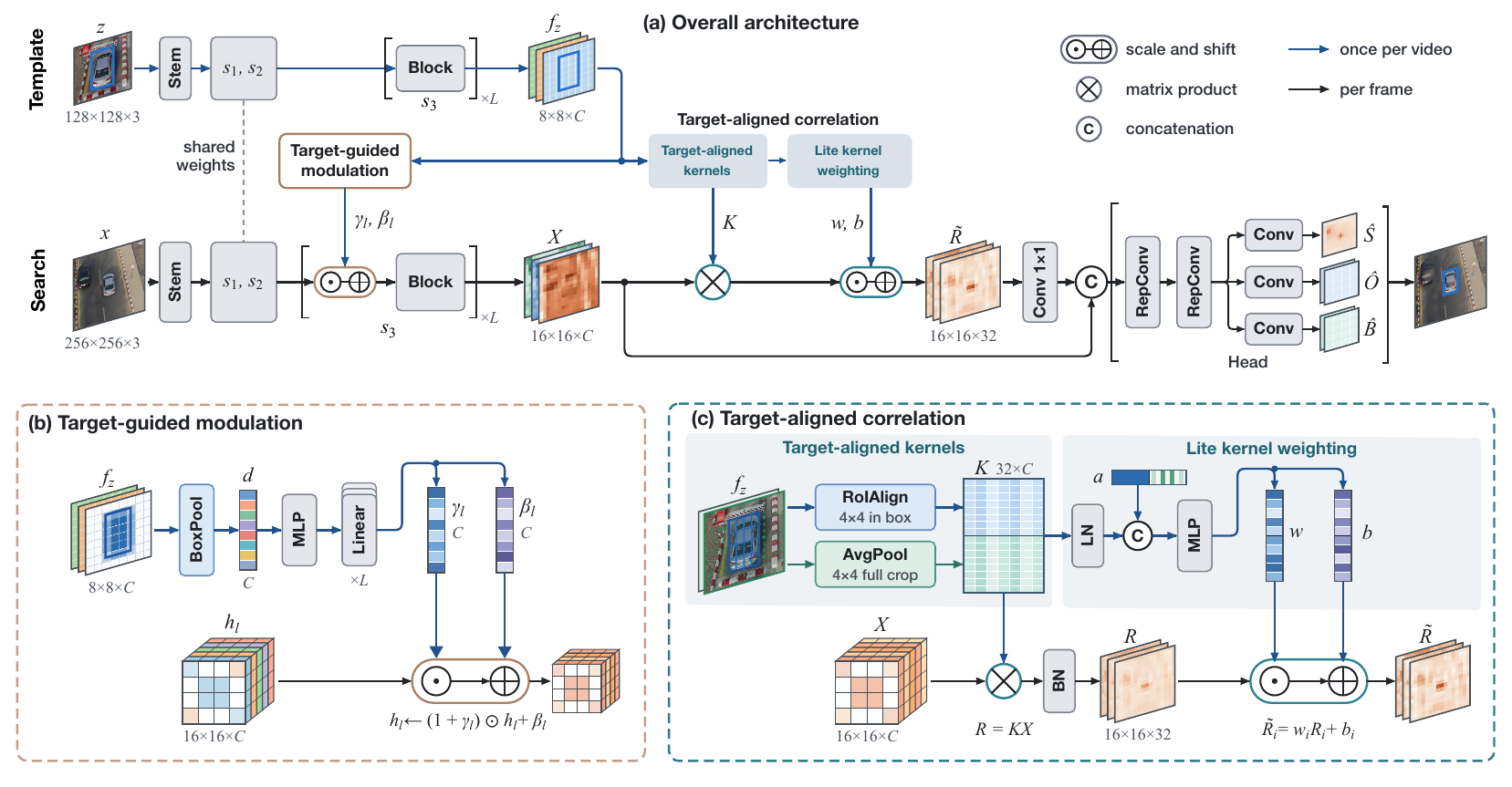}
  \caption{\textbf{\ours computes template-derived operands once per video and applies them in every frame.}
  (a) Overview.
  In the top lane, the template $z$ passes through the shared backbone $\phi$ and prepares three cached operands: the modulation parameters $(\gamma_l,\beta_l)$, the kernel bank $K$ and the kernel weights $(w,b)$.
  In the bottom lane, each search crop $x$ passes through the same backbone, the target-aligned correlation and a shared 128-channel centre head.
  The outlined multiply-add nodes (scale, then shift) mark the two places where cached operands enter the per-frame graph, and $\otimes$ is the matrix product with $K$.
  (b) Target-guided modulation pools a box-weighted descriptor $d$ from the template feature $f_z$ and predicts a scale and a shift for each modulation site.
  (c) Target-aligned kernels sample $4{\times}4$ box-aligned $\RoIAlign$ cells and $4{\times}4$ context average-pooling cells into 32 kernels $K$. Lite kernel weighting predicts, from each kernel and its fractional in-box coverage $a_i$, a gain $w_i$ and a bias $b_i$ that calibrate the normalized response $R$.
}
  \label{fig:overview}
\end{figure*}

\subsection{Rethinking correlation}
\label{sec:method_overview}
An interaction between a template and search features $X\in\mathbb{R}^{C\times HW}$ has two functions. It makes $X$ \emph{target-aware} and \emph{matches} the positions of $X$ against the target.
Trackers have used three forms of interaction,
\begin{align}
  R &= K_z X, \label{eq:form_corr}\\
  X &\leftarrow X + V_z\,\mathrm{softmax}\!\big(K_z^{\prime\top} Q_X/\sqrt{d}\big), \label{eq:form_attn}\\
  X &\leftarrow (1+\gamma_z)\odot X + \beta_z, \label{eq:form_mod}
\end{align}
where every quantity with subscript $z$ is derived from the template and the softmax runs over template tokens.
Correlation (\cref{eq:form_corr}) performs matching, where each row of $R$ is the similarity of one template kernel to every search position, but $X$ stays as the backbone extracted it.
Modulation (\cref{eq:form_mod}) makes $X$ target-aware channel by channel and compares no positions.
Attention (\cref{eq:form_attn}) does both, and attention trackers repeat it through the network, whereas a correlation tracker applies \cref{eq:form_corr} once, to the output of a backbone that never sees the target.

The forms also differ in how the template enters the per-frame graph.
In \cref{eq:form_corr,eq:form_mod} it is an operand, computed once per video, of a matrix product or of a channel-wise multiply-add.
\cref{eq:form_attn} runs, even with cached keys and values, a softmax, a token normalization and the layout operators between attention heads in every frame, which an edge NPU executes far below the throughput of convolutions and matrix products (\cref{sec:exp_rethink}).
We therefore treat correlation and modulation as complements, each covering the function the other lacks, under one design rule. We compute the template-derived parameters once per video and apply them to search features only through channel-wise multiply-adds and matrix products, without attention softmax or token normalization.

\paragraph{Overview}
\ours is a two-stream tracker whose two streams share one CNN backbone $\phi$ (\cref{fig:overview}).
The template $z$ is cropped around the target box $b$ in the first frame, and the search region $x$ in each new frame.
We keep the first three stages $s_1, s_2, s_3$ of $\phi$, at strides 4, 8 and 16, and drop the stride-32 stage, so the search feature $X$ keeps enough resolution for correlation.
Only the search stream is modulated by the target, and the template stream never sees the search region.
Let $\theta_z=\big(\{\gamma_l,\beta_l\}_{l=1}^{L},K,\{w_i,b_i\}_{i=1}^{N_k}\big)$ collect the modulation parameters of \cref{sec:method_cond} and the correlation kernels and kernel weights of \cref{sec:method_corr}, all computed from $z$ and $b$.
Tracking then splits into a once-per-video stage and a per-frame stage,
\begin{equation}
  \theta_z = g(z,b), \qquad (\hat S,\hat O,\hat B) = f(x;\theta_z),
  \label{eq:split}
\end{equation}
where $\hat S$, $\hat O$ and $\hat B$ are the score, offset and size maps.
Following the rule above, $\theta_z$ enters $f$ only as operands of channel-wise multiply-adds and of one matrix product.
\oursT, \oursS and \oursB use LowFormer-E1~\cite{nottebaum2025lowformer}, RepViT-M1.1 and RepViT-M1.5~\cite{model_repvit} as $\phi$. Everything else is shared, and the number of modulation sites follows the depth of $s_3$ (see appendix).

\subsection{Target-guided modulation}
\label{sec:method_cond}

Modulation supplies the first function, target-aware features (\cref{eq:form_mod}).
From the template feature $f_z=\phi(z)$ we pool a descriptor $d$ inside the box, averaging the cells of $f_z$ weighted by the fraction of each cell that lies inside $b$.
At each modulation site $l$ of the search stream, $d$ modulates the feature map $h_l$ channel by channel:
\begin{equation}
  h_l \leftarrow \big(1+\gamma_l(d)\big)\odot h_l + \beta_l(d),
  \label{eq:cond}
\end{equation}
where $\gamma_l$ and $\beta_l$ are predicted from $d$ by a shared MLP followed by a linear layer per site.
This is FiLM~\cite{perez2018film} with the scale written as $1+\gamma_l$. The per-site layers are initialized to zero, so training starts from a plain two-stream tracker.
We place a site before every block of $s_3$, so the target shapes the search features inside the backbone, whereas in correlation trackers such as LightFC~\cite{2024_lightfc} it first appears after the last stage.
The $s_3$ placement achieves the best trade-off between accuracy and latency among the modulated variants in \cref{sec:exp_rethink}.

Per frame, a site is one element-wise multiply-add whose operands are graph inputs.

\subsection{Target-aligned correlation}
\label{sec:method_corr}
Modulation tells the backbone what to look for, but it acts on channels and never compares search positions with the template, which is what correlation does.
Following pixel-wise correlation~\cite{yan2021alpharefine}, we use template feature vectors as $1{\times}1$ kernels and correlate them with the modulated search feature $X\in\mathbb{R}^{C\times HW}$, so the target enters the match from both sides.
Since $X$ already carries the target, we also revise what the correlation is given and what follows it.

\paragraph{Target-aligned kernels}
We sample $P{\times}P$ target cells on a grid aligned with the box using $\RoIAlign$, and average-pool $Q{\times}Q$ context cells over the whole template, giving $N_k=P^2+Q^2=32$ kernels $k_i\in\mathbb{R}^{C}$ ($P=Q=4$), scaled by $C^{-1/2}$.
LightFC~\cite{2024_lightfc} takes its kernels from a fixed $8{\times}8$ grid over the template crop, so cells mix target and background wherever the box falls.
Pairing box-aligned target kernels with global context kernels separates part-level foreground matching from surrounding background anchors, which is particularly helpful when small targets cover only a few cells on the stride-16 template grid.
Because the target box has continuous location and aspect ratio on this coarse $8{\times}8$ map, $\RoIAlign$ uses bilinear interpolation to avoid quantization misalignment and background leakage. It runs once per video in the template branch and never enters the per-frame NPU graph.
Context instead spans the fixed full-template grid and needs no box alignment, so $4{\times}4$ average pooling aggregates each $2{\times}2$ neighborhood into 16 smooth regional prototypes without resampling, avoiding single-peak dominance from max pooling and halving the kernel count relative to all 64 template cells.

\paragraph{Correlation}
The response $R=KX\in\mathbb{R}^{N_k\times HW}$ is a single matrix product per frame with $K$ as a per-video input, followed by batch normalization.

\paragraph{Lite kernel weighting}
LightFC reweights the response channels with a squeeze-and-excitation block, which is computed from the search response and therefore runs in every frame.
We instead predict the weight of each kernel from the template, once per video.
A shared MLP maps the normalized feature of $k_i$ and the fraction $a_i$ of its cell inside the box to a scalar gain $w_i$ and bias $b_i$, applied to the normalized response of that kernel:
\begin{equation}
  R_i \leftarrow w_i R_i + b_i .
  \label{eq:kcond}
\end{equation}
Target kernels have $a_i=1$, so the coverage separates target from context kernels. The last layer is initialized to $w_i=1$ and $b_i=0$.
The weighted response is reduced to 64 channels by a $1{\times}1$ convolution and concatenated with $X$ as the input of the head.
We drop the fusion layers that LightFC places at this point (depth-wise convolutions and an inverted bottleneck), since they add no accuracy once $X$ is target-aware (\cref{sec:exp_rethink}).

\subsection{Head and loss}
\label{sec:method_train}
A centre head~\cite{zhou2019centernet,2022_ostrack} predicts the score map $\hat S$, a sub-cell offset $\hat O$ and the box size $\hat B$ on the stride-16 grid $\Omega$.
Its layers are RepConv blocks~\cite{technique_ding2021repvgg}, which train with parallel branches and fold into single $3{\times}3$ convolutions for deployment.
Whereas LightFC~\cite{2024_lightfc} feeds a post-correlation fusion block into three separate 256-channel branches, we pass the concatenated search and correlation features directly into a narrower 128-channel head whose first two RepConv layers are shared across all three outputs.
Since $X$ is already target-aware and the correlation projection provides spatial evidence, sharing a compact 128-channel stem eliminates redundant per-branch convolutions and saves NPU latency without losing accuracy, while separate final layers preserve task-specific score, offset and size predictions.

We train the complete tracker with
\begin{equation}
  \mathcal{L}=\mathcal{L}_{\mathrm{focal}}+2\,\mathcal{L}_{\mathrm{GIoU}}+5\,\mathcal{L}_{1},
  \label{eq:loss}
\end{equation}
using focal loss~\cite{technique_lin2017focal} on the score map and GIoU~\cite{technique_rezatofighi2019giou} and L1 losses on the decoded box.

\begin{table*}[t]
  \centering
  \footnotesize
  \setlength{\tabcolsep}{4pt}
  \caption{\textbf{Comparison with efficient trackers on LaSOT, GOT-10k, TrackingNet and VastTrack.}
  We report per-frame MACs and graph FPS on the \rv NPU and the Raspberry Pi 5 Arm CPU.
  Competitor accuracy comes from local LaSOT/VastTrack retests and official GOT-10k/TrackingNet test-server returns.
  $^{*}$:~locally reproduced without released weights.}
  \label{tab:main}
  \begin{tabular}{l c ccc ccc ccc cc|r rr}
    \toprule
    & & \multicolumn{3}{c}{LaSOT} & \multicolumn{3}{c}{GOT-10k} & \multicolumn{3}{c}{TrackingNet} & \multicolumn{2}{c}{VastTrack} & MACs & \multicolumn{2}{c}{FPS} \\
    \cmidrule(lr){3-5} \cmidrule(lr){6-8} \cmidrule(lr){9-11} \cmidrule(lr){12-13} \cmidrule(lr){15-16}
    Tracker & Pub & AUC & P$_\mathrm{Norm}$ & P & AO & SR$_{0.5}$ & SR$_{0.75}$ & AUC & P$_\mathrm{Norm}$ & P & AUC & P & (G) & NPU & Arm \\
    \midrule
    \multicolumn{16}{l}{\textit{Correlation / CNN fusion}} \\
    LightTrack-M~\cite{2021_lighttrack} & CVPR'21 & 50.8 & 57.4 & 50.5 & 50.5 & 58.5 & 32.3 & 71.4 & 77.9 & 68.6 & 27.9 & 24.4 & 0.51 & 88.5 & 47.9 \\
    FEAR-XS~\cite{2022_fear} & ECCV'22 & 50.1 & 59.4 & 52.3 & 57.5 & 68.3 & 45.7 & 71.5 & 80.6 & 70.0 & 29.6 & 25.9 & 0.46 & 258.5 & 91.6 \\
    LightFC~\cite{2024_lightfc} & KBS'24 & 60.3 & 69.8 & 62.9 & 64.9 & 75.8 & 57.1 & 77.1 & 83.2 & 74.2 & 30.3 & 27.5 & 0.95 & 135.4 & 69.2 \\
    LightFC-vit~\cite{2024_lightfc} & KBS'24 & 63.5 & \third{74.3} & 68.0 & 67.4 & 79.2 & 59.8 & 77.8 & 84.3 & 75.5 & 30.9 & 28.6 & 2.77 & 21.5 & 11.0 \\
    \midrule
    \multicolumn{16}{l}{\textit{Attention-based interaction}} \\
    STARK-Lightning~\cite{2021_stark} & ICCV'21 & 58.1 & 66.5 & 57.8 & 59.9 & 69.9 & 48.3 & 72.7 & 77.9 & 67.4 & 28.0 & 23.3 & 1.54 & 65.8 & 38.9 \\
    HCAT-R18~\cite{2022_hcat} & ECCVW'22 & 59.3 & 68.7 & 60.9 & 65.3 & 76.7 & 56.9 & 76.6 & 82.6 & 72.9 & 29.5 & 26.0 & 2.96 & 54.1 & 27.9 \\
    HiT-Tiny~\cite{2023_hit} & ICCV'23 & 55.1 & 60.7 & 53.0 & 52.5 & 59.3 & 42.4 & 74.6 & 77.9 & 68.8 & 25.7 & 19.2 & 0.99 & 61.9 & 27.8 \\
    HiT-Small~\cite{2023_hit} & ICCV'23 & 61.1 & 68.7 & 61.7 & 62.5 & 71.2 & 54.3 & 77.8 & 81.8 & 73.2 & 28.1 & 22.8 & 1.13 & 43.9 & 20.1 \\
    HiT-Base~\cite{2023_hit} & ICCV'23 & \third{65.3} & 73.5 & 68.0 & 64.0 & 72.2 & 58.1 & 80.1 & 84.3 & 77.3 & 30.0 & 25.6 & 4.34 & 20.6 & 7.1 \\
    Aba-ViTrack~\cite{2023_abavitrack} & ICCV'23 & 64.5 & 74.2 & 68.0 & 67.1 & 77.1 & 61.8 & 79.8 & 84.9 & 76.8 & 31.4 & 27.7 & 2.86 & 16.8 & 11.4 \\
    MixFormerV2-S~\cite{2023_mixformerv2} & NeurIPS'23 & 60.9 & 69.1 & 60.7 & 61.3 & 70.4 & 51.6 & 76.4 & 81.0 & 71.1 & 29.0 & 24.9 & 4.82 & 9.5 & 7.9 \\
    SMAT~\cite{2024_smat} & WACV'24 & 61.8 & 71.3 & 64.7 & 64.5 & 74.8 & 57.9 & 78.6 & 84.2 & 75.7 & 30.8 & 27.4 & 2.49 & 20.6 & 9.8 \\
    AVTrack~\cite{2024_avtrack} & ICML'24 & 61.7 & 71.0 & 64.4 & 64.6 & 74.7 & 57.8 & 78.5 & 83.6 & 74.9 & 30.5 & 27.0 & 2.39 & 22.0 & 12.1 \\
    PromptVT~\cite{2024_promptvt} & TCSVT'24 & 64.0 & 74.0 & 67.1 & 67.6 & 78.7 & 61.2 & 78.0 & 83.4 & 74.4 & 31.2 & 28.1 & 1.93 & 30.4 & 20.8 \\
    FERMT$^{*}$~\cite{2024_fermt} & ECCV'24 & 65.0 & 74.0 & 68.1 & 68.8 & 78.9 & 64.1 & 80.8 & 85.6 & 78.1 & 31.9 & 28.8 & 2.64 & 14.1 & 10.9 \\
    AsymTrack-T~\cite{2025_asymtrack} & AAAI'25 & 61.0 & 68.8 & 61.3 & 62.4 & 71.5 & 54.8 & 76.2 & 80.9 & 71.6 & 29.4 & 24.6 & 0.64 & 52.4 & 43.3 \\
    AsymTrack-S~\cite{2025_asymtrack} & AAAI'25 & 62.6 & 71.0 & 64.4 & 65.5 & 74.8 & 58.9 & 77.9 & 82.6 & 74.0 & 30.2 & 25.9 & 0.72 & 49.5 & 36.6 \\
    AsymTrack-B~\cite{2025_asymtrack} & AAAI'25 & 64.7 & 73.1 & 67.9 & 67.7 & 76.6 & 61.5 & 79.9 & 84.4 & 77.3 & 31.2 & 27.3 & 1.77 & 19.6 & 14.9 \\
    ORTrack~\cite{2025_ortrack} & CVPR'25 & 62.9 & 72.6 & 65.9 & 65.4 & 75.3 & 58.8 & 79.2 & 84.3 & 75.9 & 30.7 & 27.2 & 2.85 & 22.1 & 12.3 \\
    ORTrack-D~\cite{2025_ortrack} & CVPR'25 & 61.3 & 70.1 & 63.3 & 63.3 & 73.2 & 56.3 & 78.4 & 83.5 & 74.6 & 30.7 & 26.9 & 1.77 & 31.7 & 22.5 \\
    SGLATrack~\cite{2025_sglatrack} & CVPR'25 & 63.1 & 72.4 & 66.0 & 66.3 & 76.2 & 60.3 & 79.5 & 84.9 & 76.5 & 31.2 & 27.7 & 1.95 & 6.2 & 19.4 \\
    FARTrack-Pico~\cite{2026_fartrack} & ICLR'26 & 57.9 & 66.6 & 59.3 & 62.6 & 72.7 & 51.0 & 75.6 & 81.4 & 70.4 & 30.4 & 25.8 & 1.43 & \textcolor{gray}{14.9} & 18.0 \\
    FARTrack-Nano~\cite{2026_fartrack} & ICLR'26 & 59.5 & 67.8 & 62.3 & 69.3 & \third{80.0} & 60.3 & 79.0 & 84.3 & 75.4 & 34.0 & 30.4 & 2.34 & \textcolor{gray}{10.4} & 9.7 \\
    FARTrack-Tiny~\cite{2026_fartrack} & ICLR'26 & 61.7 & 69.8 & 64.7 & \second{70.2} & \second{80.6} & 64.1 & 80.3 & 85.3 & 77.2 & 35.1 & 32.3 & 3.47 & \textcolor{gray}{7.5} & 6.5 \\
    UETrack-T~\cite{2026_uetrack} & CVPR'26 & 63.4 & 72.5 & 65.1 & 65.3 & 75.2 & 58.6 & 78.7 & 83.4 & 74.4 & 33.2 & 29.3 & 2.50 & 39.8 & 14.3 \\
    UETrack-S~\cite{2026_uetrack} & CVPR'26 & \first{66.5} & \first{75.7} & \second{70.3} & \first{71.1} & \first{81.3} & \first{66.9} & \second{81.4} & \third{85.9} & \second{78.6} & \second{36.6} & \third{33.7} & 3.54 & 28.9 & 9.6 \\
    MaST-Nano~\cite{2026_mast} & ECCV'26 & 59.2 & 68.7 & 60.3 & 61.6 & 72.0 & 53.8 & 77.2 & 82.9 & 72.9 & 30.7 & 26.3 & 0.68 & \textcolor{gray}{22.2} & 22.6 \\
    MaST-Tiny~\cite{2026_mast} & ECCV'26 & 64.0 & 73.1 & 67.7 & 68.0 & 77.4 & \third{64.2} & \third{81.2} & \second{86.1} & \second{78.6} & 33.2 & 29.8 & 0.96 & \textcolor{gray}{19.3} & 16.6 \\
    \midrule
    \rowcolor{oursrow} \oursT & Ours & 62.3 & 70.8 & 64.8 & 66.3 & 76.2 & 60.9 & 78.8 & 83.3 & 75.3 & 33.9 & 29.9 & 1.49 & 183.6 & 96.3 \\
    \rowcolor{oursrow} \oursS & Ours & 64.5 & 73.3 & \third{68.8} & 67.0 & 76.2 & 63.1 & 81.0 & 85.5 & \third{78.5} & \third{36.5} & \second{34.0} & 1.69 & 85.3 & 44.4 \\
    \rowcolor{oursrow} \oursB & Ours & \second{65.6} & \second{74.5} & \first{70.5} & \third{69.5} & 78.5 & \second{66.8} & \first{81.8} & \first{86.3} & \first{80.0} & \first{37.0} & \first{35.1} & 2.77 & 50.4 & 27.0 \\
    \bottomrule
  \end{tabular}
\end{table*}

\section{Experiments}
\label{sec:exp}

\subsection{Implementation details}
\label{sec:exp_setup}
\paragraph{Model}
\oursT, \oursS and \oursB use LowFormer-E1, RepViT-M1.1 and RepViT-M1.5, respectively.
All three use $128{\times}128$ templates, $256{\times}256$ search crops and stride-16 output features.
They share the correlation and head designs, with 4, 13 and 25 modulation sites in $s_3$, respectively (see \cref{tab:deploy}).

\paragraph{Training}
We initialize each backbone with ImageNet classification weights and sample 60,000 image pairs per epoch, drawing equally from the training splits of LaSOT~\cite{dataset_fan2019lasot}, GOT-10k~\cite{dataset_huang2021got10k}, COCO~\cite{dataset_lin2014coco} and VastTrack~\cite{dataset_peng2024vasttrack}.
We train for 300 epochs on four V100 GPUs with AdamW, a global batch size of 128 and the tracking loss in \cref{eq:loss}.
The initial learning rates are $8{\times}10^{-5}$ for the backbone and $8{\times}10^{-4}$ for the remaining parameters, both reduced by a factor of ten at epoch 240.

\paragraph{Inference}
We crop the template around the first-frame annotation and cache the modulation parameters, correlation kernels and kernel weights for the sequence.
At each subsequent frame, we crop the search region around the previous prediction, weight the score map with a Hann window and decode the box from the offset and size at the peak.
The template and search crop factors are 2 and 4, respectively.

\subsection{Benchmark results}
\label{sec:exp_main}
\paragraph{Evaluation protocol}
\cref{tab:main} compares trackers up to 5\,GMACs per frame. We score LaSOT and VastTrack locally and use the official GOT-10k and TrackingNet~\cite{dataset_muller2018trackingnet} servers.
We report AUC and precision, plus normalized precision on LaSOT and TrackingNet. GOT-10k uses AO and success rates at 0.50 and 0.75 IoU.
All three \ours variants use their mixed-training epoch-300 checkpoints.

\paragraph{Tracking accuracy}
\cref{tab:main} shows that \ours delivers strong accuracy across all four benchmarks while keeping edge latency low.
On the long-term LaSOT benchmark, AUC rises from 62.3 to 64.5 to 65.6 across \oursT, \oursS and \oursB, with each variant ahead of its counterpart in the HiT and AsymTrack families.
Our lightest variant, \oursT, already outperforms the correlation tracker LightFC by 2.0 AUC points (62.3 vs.\ 60.3) while running faster on both the NPU (183.6 vs.\ 135.4\,FPS) and Arm CPU (96.3 vs.\ 69.2\,FPS), and also surpasses heavier attention trackers such as MixFormerV2-S and FARTrack-Tiny.
On VastTrack, which tests generalization across a much wider set of object categories, \oursB leads the entire table with 37.0 AUC and 35.1 precision, followed closely by \oursS at 36.5 AUC and 34.0 precision (ranking second in precision ahead of UETrack-S).
On the large-scale TrackingNet benchmark, \oursB ranks first across all three metrics (81.8 AUC, 86.3 P$_\mathrm{Norm}$ and 80.0 P), while \oursT (78.8 AUC) UETrack-T while being several times faster.
On the GOT-10k, \oursB achieves 69.5 AO and ranks second in SR$_{0.75}$ (66.8), while \oursT (66.3 AO) again beats UETrack-T.
For a per-attribute breakdown on LaSOT, please refer to the appendix.

\paragraph{Edge runtime evaluation}
We measure per-frame network graph throughput on the Rockchip \rv NPU (RKNN 2.3.2, FP16, 950\,MHz) and the Raspberry Pi 5 Arm CPU (ncnn, four Cortex-A76 threads).
Timed calls use cached template operands where supported, and gray NPU values fail the numerical check (see appendix C for more details).
All three \ours per-frame graphs run in FP16 on both devices, entirely on the NPU for RKNN.
On the NPU, \oursT runs $3.0\times$ as fast as HiT-Tiny and $3.5\times$ as fast as AsymTrack-T, while \oursB runs $2.3\times$ as fast as LightFC-vit at the same reported MAC count.
On the Pi~5, \oursT is $1.4\times$ as fast as LightFC, and \oursB is the fastest tracker among those reaching at least 64 LaSOT AUC.
One-time template-branch latencies are in \cref{tab:deploy}.
Nine attention-based competitor variants, but no correlation tracker, fail the FP16 check on at least one device, and the traceable failures arise in a softmax or token selection (see appendix C).

\paragraph{Other benchmarks}
\begin{table}[t]
  \centering
  \scriptsize
  \setlength{\tabcolsep}{4pt}
  \caption{\textbf{Results on LaSOT$_\mathrm{ext}$, TNL2K, NfS and UAV123.}
  Competitor results are local retests of the released checkpoints. \first{Red} and \second{blue} mark the best and second-best value of each column.}
  \label{tab:other_general}
  \begin{tabular}{l cc cc cc cc}
    \toprule
    & \multicolumn{2}{c}{LaSOT$_\mathrm{ext}$} & \multicolumn{2}{c}{TNL2K} & \multicolumn{2}{c}{NfS} & \multicolumn{2}{c}{UAV123} \\
    \cmidrule(lr){2-3} \cmidrule(lr){4-5} \cmidrule(lr){6-7} \cmidrule(lr){8-9}
    Tracker & AUC & P & AUC & P & AUC & P & AUC & P \\
    \midrule
    LightFC~\cite{2024_lightfc} & 41.4 & 45.4 & 49.6 & 48.1 & 63.7 & 76.7 & 64.7 & 85.1 \\
    HiT-Tiny~\cite{2023_hit} & 35.7 & 36.0 & 41.3 & 33.8 & 53.4 & 60.4 & 59.7 & 76.8 \\
    MixFormerV2-S~\cite{2023_mixformerv2} & 43.1 & 45.7 & 45.9 & 40.1 & 62.6 & 74.1 & 65.5 & 86.4 \\
    SMAT~\cite{2024_smat} & 40.7 & 43.9 & 50.9 & 48.7 & 63.4 & 75.8 & 64.8 & 84.7 \\
    PromptVT~\cite{2024_promptvt} & \first{45.3} & \first{49.4} & 51.1 & 49.2 & 64.9 & 78.2 & 66.6 & \first{87.5} \\
    AsymTrack-T~\cite{2025_asymtrack} & 42.7 & 45.9 & 48.6 & 44.5 & 63.3 & 74.5 & 64.4 & 84.2 \\
    ORTrack-D~\cite{2025_ortrack} & 42.5 & 46.3 & 49.5 & 46.2 & 64.3 & 76.5 & 65.7 & 86.0 \\
    FARTrack-Pico~\cite{2026_fartrack} & 39.8 & 41.9 & 48.2 & 45.0 & 62.3 & 76.2 & 63.0 & 84.2 \\
    UETrack-T~\cite{2026_uetrack} & 42.1 & 46.2 & 45.9 & 40.8 & 63.5 & 75.9 & \first{66.9} & 86.2 \\
    MaST-Nano~\cite{2026_mast} & 41.1 & 43.3 & 47.2 & 42.3 & 61.9 & 72.8 & 62.4 & 81.0 \\
    AVTrack~\cite{2024_avtrack} & 42.5 & 46.6 & 49.4 & 46.5 & 63.5 & 75.1 & 66.1 & 86.4 \\
    SGLATrack~\cite{2025_sglatrack} & 42.5 & 46.2 & 51.4 & 49.3 & 64.7 & 76.9 & \second{66.8} & \second{87.4} \\
    \midrule
    \rowcolor{oursrow} \oursT & 41.3 & 45.4 & 50.8 & 48.5 & \second{65.7} & \second{78.7} & 64.3 & 83.5 \\
    \rowcolor{oursrow} \oursS & 43.2 & \second{47.9} & \second{53.1} & \second{52.8} & \second{65.7} & 78.4 & 65.0 & 83.7 \\
    \rowcolor{oursrow} \oursB & \second{44.5} & \first{49.4} & \first{54.0} & \first{54.5} & \first{66.4} & \first{79.6} & 64.8 & 83.9 \\
    \bottomrule
  \end{tabular}
\end{table}

\cref{tab:other_general} reports one-pass evaluation on four more benchmarks. \oursB leads on TNL2K and NfS with 54.0 and 66.4 AUC, respectively.

\subsection{Controlled study on rethinking correlation}
\label{sec:exp_rethink}

\label{sec:exp_interaction}\label{sec:exp_ablation}
\begin{table}[t]
  \centering
  \footnotesize
  \caption{\textbf{Component ablation and attention alternatives on \oursT.}
  The top block removes one component at a time from the final \oursT, and the bottom block replaces correlation with a cross-attention read or replaces both modulation and correlation with joint attention.
  Latency uses $256$-pixel search crops. }
  \label{tab:rethink_ablation}
  \setlength{\tabcolsep}{2.2pt}
  \begin{tabular}{l r r r r r r}
    \toprule
    & \multicolumn{2}{c}{LaSOT} & \multicolumn{2}{c}{VastTrack} & \multicolumn{2}{c}{ms} \\
    \cmidrule(lr){2-3} \cmidrule(lr){4-5} \cmidrule(lr){6-7}
    Variant & AUC & $\Delta$ & AUC & $\Delta$ & NPU & Arm \\
    \midrule
    \rowcolor{oursrow}\textbf{\oursT} & 61.55 & --- & 33.46 & --- & 5.44 & 10.34 \\
    w/o modulation       & 60.96 & $-0.59$ & 32.69 & $-0.77$ & 5.32 & 10.23 \\
    w/o correlation      & 59.07 & $-2.48$ & 32.28 & $-1.18$ & 5.24 & 9.66 \\
    w/o target alignment & 60.82 & $-0.73$ & 33.22 & $-0.24$ & 5.44 & 10.39 \\
    w/o kernel weighting & 60.93 & $-0.62$ & 33.35 & $-0.11$ & 5.43 & 10.35 \\
    \midrule
    corr.\ $\to$ cross attn.  & 58.03 & $-3.52$ & 26.85 & $-6.61$ & 7.85 & 23.96 \\
    corr.\ $\to$ joint attn.  & 58.61 & $-2.94$ & 26.37 & $-7.09$ & 9.02 & 29.15 \\
    \bottomrule
  \end{tabular}
\end{table}

\begin{table}[t]
  \centering
  \footnotesize
  \caption{\textbf{Modulation stages.} We vary the stages that receive modulation, with target-aligned correlation fixed.}
  \label{tab:rethink_sites}
  \setlength{\tabcolsep}{5pt}
  \begin{tabular}{c c c r r r r r r}
    \toprule
    \multicolumn{3}{c}{Stages} & \multicolumn{2}{c}{LaSOT} & \multicolumn{2}{c}{VastTrack} & \multicolumn{2}{c}{ms} \\
    \cmidrule(lr){1-3} \cmidrule(lr){4-5} \cmidrule(lr){6-7} \cmidrule(lr){8-9}
    $s_1$ & $s_2$ & $s_3$ & AUC & P & AUC & P & NPU & Arm \\
    \midrule
    \cmark & \cmark & \cmark     & \textbf{61.67} & 63.85 & {33.35} & 29.35 & 5.92 & 11.08 \\
    \cmark &        &            & 61.20 & 63.87 & \textbf{33.51} & 29.61 & 5.73 & 10.85 \\
           & \cmark &            & 60.49 & 62.74 & 32.56 & 29.31 & 5.42 & 10.36 \\
    \rowcolor{oursrow}& & \cmark & 61.55 & \textbf{64.20} & 33.46 & \textbf{29.66} & {5.44} & {10.34} \\
    \bottomrule
  \end{tabular}
\end{table}

\begin{table}[t]
  \centering
  \footnotesize
  \setlength{\tabcolsep}{5pt}
  \caption{\textbf{Correlation design.} We report LaSOT accuracy alongside per-frame latency.
  Grid denote 64 fixed grid kernels.
  SE and \na denote per-frame squeeze-and-excitationvand no weighting.
  Fusion indicates LightFC's post-correlation fusion layers.
  }
  \label{tab:rethink_correlation}
  \begin{tabular}{l l c r r r r}
    \toprule
    & & & \multicolumn{2}{c}{LaSOT} & \multicolumn{2}{c}{ms} \\
    \cmidrule(lr){4-5} \cmidrule(lr){6-7}
    Kernels & Weighting & Fusion & AUC & P & NPU & Arm \\
    \midrule
    Grid & SE  & \cmark & {61.50} & \textbf{64.41} & 6.01 & 11.00 \\
    Grid & LKW & \cmark & 61.11 & 64.26 & 5.68 & 10.45 \\
    TA   & SE  & \cmark & 60.83 & 63.72 & 5.72 & 10.51 \\
    TA   & LKW & \cmark & 61.28 & 63.26 & 5.53 & 10.39 \\
    \rowcolor{oursrow}TA   & LKW & \xmark & \textbf{61.55} & 64.20 & 5.44 & 10.34 \\
    \bottomrule
  \end{tabular}
\end{table}

\cref{tab:rethink_ablation,tab:rethink_sites,tab:rethink_correlation} report averages over two seeds (200 epochs, $128{\times}128$ template, $256{\times}256$ search) and latency over 256 crops on \rv NPU and Pi 5 Arm CPU.
Without target alignment, the 16 target kernels revert to a fixed template grid.

\paragraph{Is correlation needed, and what does it cost?}
Explicit correlation is the largest single accuracy factor and nearly free on edge hardware (\cref{tab:rethink_ablation}).
Removing it costs 2.48 LaSOT and 1.18 VastTrack AUC but saves only 0.20\,ms on the NPU and 0.68\,ms on Arm.
Cross and joint attention in its place lose 2.9--3.5 LaSOT and 6.6--7.1 VastTrack AUC at 1.4--1.7$\times$ the NPU and 2.3--2.8$\times$ the Arm latency.
The VastTrack loss is about twice the LaSOT loss, suggesting that a learned attention readout generalizes worse than the built-in 2D correspondence of correlation to VastTrack's wider range of classes.

\paragraph{What does correlation lack, and where to modulate?}
Removing target-guided modulation costs 0.59 LaSOT and 0.77 VastTrack AUC, while the modulation takes only 0.12\,ms on the NPU and 0.11\,ms on Arm, under 2.3\% of frame time (\cref{tab:rethink_ablation}).
Among placements (\cref{tab:rethink_sites}), modulating all three stages gains only 0.12 LaSOT AUC over $s_3$ alone, trails it in LaSOT precision and on VastTrack, and adds 0.48\,ms on the NPU and 0.74\,ms on Arm on $16\times$ and $4\times$ larger early maps.
Modulating only before the four stride-16 $s_3$ blocks gives the best precision on both benchmarks at nearly the lowest latency.
A channel-wise scale selects whole channels, which we expect to matter most where channels encode object parts rather than low-level edges.

\paragraph{How should correlation adapt to target-aware features?}
Because its operands are cached once per video, the upgraded correlation costs almost nothing per frame (\cref{tab:rethink_ablation}). Target alignment adds 0.73 LaSOT and 0.24 VastTrack AUC at the same NPU latency, and lite kernel weighting adds 0.62 and 0.11 for 0.01\,ms.
In \cref{tab:rethink_correlation}, target alignment helps only with lite kernel weighting and lowers AUC with SE (61.50 to 60.83), likely because SE, computed from the search response, cannot tell context kernels from target kernels, whereas lite kernel weighting receives each kernel's box coverage $a_i$.
Dropping the fusion layers then gives the highest AUC in the table (61.55) at the lowest latency, 0.57\,ms (9.5\%) faster than LightFC's correlation on the NPU and 0.66\,ms faster on Arm.

\begin{figure*}[t]
  \centering
  \includegraphics[width=\linewidth]{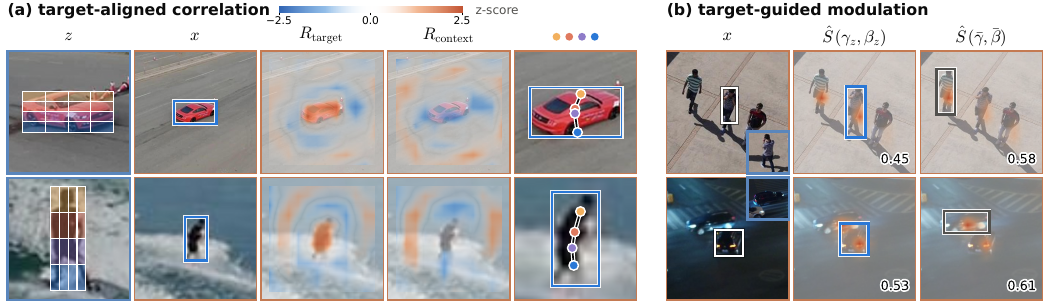}
  \caption{\textbf{Visualizing target-aligned correlation and target-guided modulation in \oursT.}
  (a) We show the template $z$ with a $4{\times}4$ target-kernel grid, the search $x$ with the blue prediction, the mean standardized responses of 16 target and 16 context kernels (red and blue denote above and below the spatial mean), and the positive-response centroids of kernel rows connected in matching colors from top to bottom.
  (b) Search $x$ with white ground truth and template inset, followed by scores and predictions using target operands $(\gamma_z,\beta_z)$ (blue) or their cross-video mean $(\bar\gamma,\bar\beta)$ (gray), with all weights and correlation kernels fixed.}
  \label{fig:mechanism}
\end{figure*}
\begin{figure}[t]
  \centering
  \includegraphics[width=\linewidth]{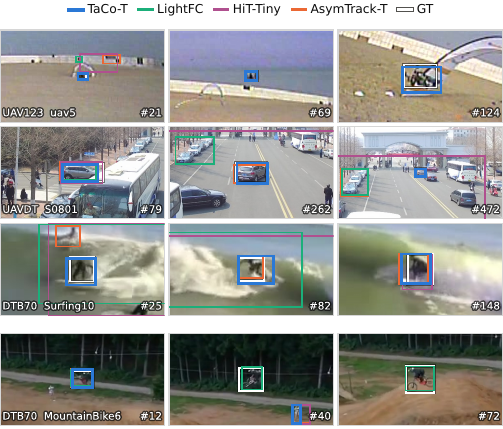}
  \caption{\textbf{Qualitative comparison.} Frames are sampled at 15, 50 and 90\% of each sequence.
  In rows 1--3, \oursT stays on a target under 20 pixels wide, among similar vehicles and in breaking surf.}
  \label{fig:qualitative}
\end{figure}

\paragraph{Visualizing correlation and modulation}
\cref{fig:mechanism} visualizes both interactions from measured maps.
In \cref{fig:mechanism}a, $R_{\mathrm{target}}$ activates tightly inside the predicted box and suppresses background, whereas $R_{\mathrm{context}}$ responds complementarily over the surround. The positive-response centroids of the four $4{\times}4$ grid rows preserve their top-to-bottom order on both elongated and upright targets.
In \cref{fig:mechanism}b, the target's operands $(\gamma_z,\beta_z)$ keep $\hat{S}$ peaked on the target despite nearby distractors, whereas substituting the cross-video mean $(\bar\gamma,\bar\beta)$ with identical kernels shifts the peak onto a distractor and pulls the box off target, confirming that modulation supplies instance selectivity while correlation supplies ordered localization (\cref{sec:method_overview}).
\Cref{fig:qualitative} shows the same behavior on UAV scnarios where competing trackers drift or lose scale, along with a failure case in which \oursT switches to a nearby bystander (with more in appendix).

\subsection{Same-backbone comparison across trackers}
\label{sec:exp_same_backbone}

\begin{table}[t]
  \centering
  \footnotesize
  \setlength{\tabcolsep}{2.5pt}
  \newcommand{\tabdiff}[1]{\,{\scriptsize(#1)}}
  \newcommand{\posdiff}[1]{\,{\scriptsize\third{(#1)}}}
  \newcommand{\neudiff}[1]{\,{\scriptsize\textcolor{gray}{(#1)}}}
  \caption{\textbf{Transfer across backbones on LaSOT.}
  We report training epochs (Ep.), LaSOT AUC and \rv NPU latency in ms, with changes from the original tracker in parentheses.}
  \label{tab:same_backbone}
  \begin{tabular}{l rrr rrr}
    \toprule
    & \multicolumn{3}{c}{Original} & \multicolumn{3}{c}{\ours} \\
    \cmidrule(lr){2-4} \cmidrule(lr){5-7}
    Source & Ep. & AUC$\uparrow$ & ms$\downarrow$ & Ep. & AUC$\uparrow$ & ms$\downarrow$ \\
    \midrule
    HiT-Tiny   & 1500 & 55.1 & 16.16 & 300 & \best{61.2}\posdiff{+6.1} & \best{13.49}\posdiff{$-$2.68} \\
    HiT-Small  & 1500 & 61.1 & 22.79 & 300 & \best{63.1}\posdiff{+2.0} & \best{18.60}\posdiff{$-$4.19} \\
    LightFC   & 400  & 60.3 & 7.39  & 300 & \best{61.8}\posdiff{+1.5}       & 7.24\neudiff{$-$0.15} \\
    AsymTrack-T             & 500  & 61.0 & 19.08 & 300 & 61.1\neudiff{+0.1}       & \best{12.54}\posdiff{$-$6.54} \\
    \bottomrule
  \end{tabular}
  \vspace{-0em}
\end{table}

To separate the gain of our interaction design from backbone choice, we keep the published backbones of three distinct tracking paradigms fixed and replace only their native interaction with modulation and correlation(\cref{tab:same_backbone}).
Across all three families, the replacement improves accuracy and reduces \rv NPU latency simultaneously with fewer training epochs.
For \textbf{HiT}~\cite{2023_hit}, which joins template and search tokens inside its LeViT backbones, the replacement gives the largest gains, but the attention and layout operators left in LeViT still keep the transferred tracker over $2.4\times$ slower than our pure-CNN \oursT (13.49 vs.\ 5.45\,ms).
For \textbf{LightFC}~\cite{2024_lightfc}, a pure CNN tracker built around pixel-wise correlation and convolutional fusion, the 1.5-point gain without added NPU cost shows that target-aware backbone features also benefit an external convolutional backbone.
For \textbf{AsymTrack-T}~\cite{2025_asymtrack}, which already caches the template through prototype-based modulation, the final \ours interaction matches its accuracy at 34\% lower NPU latency, showing that how cached operands are applied per frame governs edge throughput.

\vspace{-1em}

\section{Conclusion}
\label{sec:conclusion}

We set out to rethink correlation for edge tracking, and on-device measurements change what should be asked of it.
On edge NPUs and Arm CPUs, explicit correlation is nearly free and essential for accuracy, whereas attention's softmax, normalization, and layout ops incur heavy latency concealed by MAC counts.
What correlation lacked was a target-aware search backbone, supplied by cached channel-wise scale and shift at a fraction of attention's cost.
\ours realizes this division via target-guided modulation and target-aligned correlation with cached per-video operands, transferring across backbones and reaching 184\,FPS at 62.3\% LaSOT AUC and 50\,FPS at 65.6\% on \rv (96 and 27\,FPS on Arm).
These findings target edge NPUs and Arm CPUs under a fixed template. Adapting the operands online remains future work.

\clearpage
{
    \small
    
\section*{LLM Usage Statement}
We used LLM-based coding agents (Deepseek Harness and Copilot in VS Code) throughout this work.
\textbf{Code and experiments.}
The agents are used for code review, debugging, training configurations checking and evaluation lauching on our GPU servers, and compiling the result tables from the run logs.
\textbf{Deployment and latency measurement.}
We use coding agents to export our and collected tracker models to ONNX, converted them with RKNN-Toolkit and ncnn runtime, and ran the on-device benchmarks under a unified protocol.
Every reported accuracy and latency number comes from an actual run whose logs are kept. Before timing, the on-device outputs of every deployed graph were checked against a PyTorch Runtime FP32 reference.
\textbf{Writing.}
The agents helped polish the text, LaTeX tables, plotting scripts and checking bibliography format. We revised all text and checked every citation against its original source in person.

    \bibliographystyle{ieeenat_fullname}
    \bibliography{main}
}

\clearpage
\clearpage
\renewcommand{\thepage}{S\arabic{page}}
\setcounter{page}{1}
\maketitlesupplementary
\appendix
\begin{suppbib}

\section{Architecture details}
\label{sec:supp_arch}
All three models use a shared template/search backbone and stride-16 output features.
The template and search inputs are $128\times128$ and $256\times256$, producing $8\times8$ and $16\times16$ terminal grids.
The stride-32 stage is omitted in the three deployed CNN variants.
A box-pooled template descriptor generates a channel-wise scale and shift before each block of $s_3$.
There are 4, 13 and 25 modulation sites in \oursT, \oursS and \oursB, respectively.
The template branch is evaluated without these modulations.

\begin{table}[b]
  \centering
  \footnotesize
  \setlength{\tabcolsep}{1.5pt}
  \caption{\textbf{TaCo configurations and graph latency.}
  Params count the deployed template and per-frame graphs. Raspberry Pi 5 uses four Cortex-A76 threads.
  On the \rv, the per-frame graph runs entirely on the NPU. Template initialization uses RKNN FP16 for the backbone and kernel generation. The CPU computes box geometry and FP32 modulation parameters.}
  \label{tab:deploy}
  \begin{tabular}{l c c c}
    \toprule
    & \oursT & \oursS & \oursB \\
    \midrule
    Body & LowFormer-E1 & RepViT-M1.1 & RepViT-M1.5 \\
    Params (M) & 5.53 & 6.74 & 11.89 \\
    GMACs / frame & 1.49 & 1.69 & 2.77 \\
    Modulation sites & 4 & 13 & 25 \\
    \midrule
    \multicolumn{4}{l}{\textbf{Per-frame latency (ms)}} \\
    NPU, RKNN FP16 & 5.45 & 11.72 & 19.86 \\
    Pi 5, ncnn FP16 & 10.38 & 22.52 & 37.00 \\
    \midrule
    \multicolumn{4}{l}{\textbf{Template-branch latency (ms, once / video)}} \\
    \rv, NPU + CPU & 19.41 & 24.31 & 30.36 \\
    Pi 5, ncnn FP16 & 5.45 & 7.27 & 11.98 \\
    \bottomrule
  \end{tabular}
\end{table}

The correlation cache contains 16 kernels pooled inside the target box and 16 pooled over the template crop.
Kernel weighting uses the feature of each kernel and the fractional overlap of its cell with the target box.
The correlation response is normalized, weighted kernel by kernel and reduced to 64 channels by a $1{\times}1$ convolution before concatenation with the search feature. The fusion layers of LightFC are not used.
The 128-channel RepConv centre head shares its first two layers between the score, offset and size branches.
Parameter counts, per-frame MACs, and both per-frame and one-time template-branch latencies are in \cref{tab:deploy}.

\paragraph{Batch normalization and kernel weighting}
During inference, the correlation branch uses the running mean and variance accumulated during training, along with the learned BatchNorm affine parameters.
For each response channel $i$, the order is
\begin{equation}
  \widetilde R_i = w_i\,\BN^{\mathrm{eval}}_i\!\bigl((KX)_i\bigr)+b_i .
  \label{eq:supp_bn_kcond}
\end{equation}
The BatchNorm parameters are fixed across videos. The gain $w_i$ and bias $b_i$ are computed from the first-frame template and adjust the normalized response for that target at every frame, without updating the running statistics.

\section{Training and evaluation details}
\label{sec:supp_train}
\begin{table}[t]
  \centering\footnotesize
  \caption{\textbf{Training settings.} All models use 300 epochs and tracking supervision unless noted otherwise.}
  \label{tab:supp_train}
  \setlength{\tabcolsep}{3pt}
  \begin{tabular}{p{0.49\linewidth} p{0.43\linewidth}}
    \toprule
    Optimizer & AdamW \\
    Learning rate / backbone multiplier & $8\times10^{-4}$ / 0.1 \\
    Weight decay / gradient clipping & $10^{-4}$ / 0.1 \\
    Global batch size & 128 \\
    Pairs per epoch (nominal) & 60,000 \\
    Epochs & 300 \\
    Learning-rate step & epoch 240, factor 0.1 \\
    Template / search crop & $128^2$ / $256^2$ \\
    Template / search context factor & 2 / 4 \\
    Precision & FP16 AMP \\
    Dataset sampling & Equal across the four datasets \\
    \bottomrule
  \end{tabular}
\end{table}
Backbones start from their classification initialization.
All three models are trained on four V100 GPUs with 32 samples per GPU and seed 42.
The same global batch size and optimizer settings apply across the family (\cref{tab:supp_train}).

The controlled study of \cref{sec:exp_rethink} uses seeds 42 and 43 and 200 epochs with the learning-rate step at epoch 160. The optimizer, sampling and global batch size of 128 follow \cref{tab:supp_train}.
Evaluation in \cref{sec:exp_rethink} uses the full 280-sequence LaSOT test split and the 3,500-sequence VastTrack test split.

GOT-10k test and TrackingNet test contain 180 and 511 sequences in our evaluation, with 22,834 and 225,589 prediction rows.
Their reported scores come from the official servers, with complete coverage.
VastTrack uses its 3,500-sequence test split.
For the main comparison, we evaluate all three adopted checkpoints with native PyTorch FP16 autocast on a V100 across all four benchmarks.

\paragraph{GOT-10k-only training}
For \cref{tab:supp_got_only}, we train each variant for 100 epochs on all 9,335 official GOT-10k training sequences, with the learning-rate drop after epoch 80, and compare against trackers reporting official GOT-10k-only scores.
\begin{table}[ht]
  \centering
  \footnotesize
  \setlength{\tabcolsep}{3.4pt}
  \caption{\textbf{GOT-10k test with GOT-10k-only training.} Official test scores (\%) under the one-shot protocol (trained solely on the GOT-10k training split), alongside Raspberry Pi~5 four-thread ncnn FPS measured under the \cref{tab:main} protocol. Accuracy is from the respective papers.}
  \label{tab:supp_got_only}
  \begin{tabular}{l c rrr|r}
    \toprule
    Tracker & Pub & AO & SR$_{0.5}$ & SR$_{0.75}$ & Arm FPS \\
    \midrule
    TransT~\cite{2021_transt} & CVPR'21 & 67.1 & 76.8 & 60.9 & 0.8 \\
    STARK-ST50~\cite{2021_stark} & ICCV'21 & 68.0 & 77.7 & 62.3 & 2.0 \\
    OSTrack-256~\cite{2022_ostrack} & ECCV'22 & 71.0 & 80.4 & 68.2 & 1.7 \\
    LiteTrack-B4~\cite{2024_litetrack} & ICRA'24 & 65.2 & 74.7 & 57.7 & 6.5 \\
    FERMT~\cite{2024_fermt} & ECCV'24 & 69.6 & 80.1 & 63.2 & 10.9 \\
    MST~\cite{2025_multistate_tracker} & MM'25 & 69.6 & 79.8 & 64.1 & 3.6 \\
    \midrule
    \rowcolor{oursrow}\oursT & Ours & 61.8 & 70.9 & 54.5 & 96.3 \\
    \rowcolor{oursrow}\oursS & Ours & 65.6 & 75.1 & 61.1 & 44.4 \\
    \rowcolor{oursrow}\oursB & Ours & 66.7 & 76.5 & 62.4 & 27.0 \\
    \bottomrule
  \end{tabular}
\end{table}

\section{Deployment details}
\label{sec:supp_deploy}
The \rv measurements use RKNN Toolkit/runtime 2.3.2 and a 950\,MHz NPU clock.
For per-frame latency, the C benchmark times \texttt{rknn\_run} with five warm-ups, 1,000 calls per round, and five interleaved rounds with a fixed anchor.
We report the median of the round means.
CPU deep idle is disabled temporarily for the unified family and key-comparator measurements and restored after the session, and the board is locked against concurrent benchmarks.
Input upload, output retrieval, crop preprocessing, template initialization and host box decoding are outside the timed call.
The per-frame CPU results use a Raspberry Pi 5 (four Cortex-A76 cores up to 2.4\,GHz, kernel 6.12 with NUMA interleaving) and ncnn 1.0.20261020 with four threads bound to cores 0--3 and a shared runner. Trackers added after the main session, including the adopted \ours models, were timed in later sessions of the same runner whose anchors reproduce the earlier values within 2\%.
Each graph is first timed with its exported settings and then with the same six runtime settings (Winograd and SGEMM variants), and the fastest setting that passes the numerical check is kept.
FP16 is preferred, and a graph whose FP16 outputs fail the check falls back to validated partial or full FP32 ncnn precision.
The check requires relative L2 error at most 0.02 and maximum absolute error at most $\max(0.005,\,0.05\times$ the reference peak$)$ against the FP32 reference.
Each timing takes three groups of about 2.6\,s (10--250 calls) and reports the median of the group means. The timed call covers host input to host output with each architecture's template caching.
Firmware throttling and under-voltage flags are sampled every second, and any flagged timing is repeated.
The \ours CPU graphs are converted from the adopted checkpoints and pass the check in FP16.

The template graph runs once per sequence on the $128{\times}128$ template crop to produce the modulation operands, the box-aligned $\RoIAlign$ and context average-pooled kernels, and the lite kernel weights (\cref{tab:deploy}).
Template timing starts from the normalized crop and target box and ends when all cache tensors are available on the host.
We report the median of five round means using eight LaSOT first-frame crops.
The maximum relative L2 error of the template caches against ONNX Runtime FP32 is 0.011 across both devices.

RepConv branches and RepViT blocks are fused before RKNN conversion.
The per-frame graph returns score logits, offsets and sizes. The score sigmoid and final decoding run on the host.
To avoid response overflow, the correlation input is scaled by an exact power of two and the compensating scale is folded into the following normalization.
All adopted per-frame graphs execute entirely on the NPU.

For each adopted checkpoint, eight real crop pairs are checked against ONNX Runtime FP32.
The acceptance gate requires relative L2 error at most 0.02, identical score-peak cells, and decoded-box IoU at least 0.98.
All three adopted checkpoints pass, with relative L2 error at most 0.009 and decoded-box IoU at least 0.997.

\paragraph{Numerical check outcomes}
Before accepting an FP16 graph in \cref{tab:main}, we verify every exported output against its ONNX Runtime FP32 reference.
On both devices, the predeclared numerical gate requires relative $L_2\leq0.02$ and maximum absolute error at most $\max(0.005,\,0.05\lVert Y_{\mathrm{ref}}\rVert_\infty)$ (with exact agreement on integer and boolean outputs).
On the \rv NPU, passing graphs must also run with zero CPU compute operators; for \ours, we additionally verify identical score-peak cells (after sigmoid and Hann weighting) and decoded-box IoU at least 0.98 across eight real crop pairs.
Failed graphs on the \rv have their measured FP16 FPS displayed in gray, while failed graphs on the Raspberry Pi~5 fall back to their validated partial or full FP32 ncnn deployment.

Across the 31 models in \cref{tab:main}, all 4/4 correlation/CNN-fusion baselines and all 3/3 \ours per-frame graphs pass in FP16 on both devices (including LightFC-vit with its ViT backbone).
Among the 24 attention-based variants, 19/24 pass on the \rv and 16/24 pass on the Pi~5; nine attention variants fail on at least one device.
On the \rv NPU, the three FARTrack and two MaST graphs fail and are shown in gray.
In FARTrack, the top-1 score of the softmax over 605 coordinate bins is about 0.035 below the FP32 reference, which we attribute to FP16 flooring of the many negligible exponentials, and the argsort that selects template tokens for later frames flips up to 6 of 49 tokens near the pruning boundary (the decoded box still agrees within 0.006 of the search crop, so gray marks a failure under the predeclared rule rather than a demonstrated tracking failure).
In MaST-Nano and MaST-Tiny, FP16 rounding perturbs the top-$k$ token ordering, and in MaST-Nano also alters the selected token set on one test case.
On the Pi~5, HiT-Tiny/Small/Base, SMAT, MaST-Nano/Tiny and FARTrack-Nano/Tiny pass only in ncnn FP32, and MaST also runs its top-$k$ token selection on the CPU; their Arm FPS therefore includes FP32 arithmetic.

\paragraph{Throughput, memory and power during steady inference}
\Cref{tab:supp_runtime_resources} reports batch-one graph execution with fixed input tensors across the seven multi-variant families and FEAR-XS from \cref{tab:main}. For models with template caching, the prepared operands are supplied to the per-frame graph, and the \ours rows use the submitted epoch-300 weights.
For \rv and Raspberry Pi~5, \cref{tab:supp_runtime_resources} carries over the graph FPS from \cref{tab:main}, whereas Orin Nano FPS counts completed TensorRT graph calls per second with device-resident I/O.
The \texttt{Mem} column reports the primary runtime memory footprint on each platform: TensorRT-managed GPU memory allocation after engine and context creation on Orin Nano, RKNN DMA allocation reported by \texttt{RKNN\_QUERY\_MEM\_SIZE} after warm-up on the \rv, and inference-process peak resident set size (RSS, including the ncnn runner) on the Raspberry Pi~5.

Orin Nano runs TensorRT 10.3.0 in MAXN\_SUPER mode with module input power (\texttt{VDD\_IN}) sampled over 12-second inference trials (idle power 5.34\,W). The three TensorRT \ours engines pass verification on a real crop pair against ONNX Runtime FP32 (maximum relative $L_2$ of 0.0026, matching score peaks and decoded-box IoU $\geq0.998$). Competitor rows reuse their archived verified TensorRT engines; when an FP16 engine fails the FP32 numerical gate (HiT-Tiny, FARTrack-Nano/Tiny and MaST-Nano/Tiny), the verified FP32 engine is reported and marked with .
Raspberry Pi~5 memory and power are measured with four ncnn threads on the same validated FP16/FP32 graphs as \cref{tab:main} ( denotes FP32) under a 1.8\,GHz CPU frequency cap so that the 12-second power trials remain free of undervoltage and throttling; power sums the monitored PMIC supply rails excluding HDMI (idle power 2.29\,W).
The \rv runs RKNN 2.3.2 at 950\,MHz, staging each model on eMMC and querying \texttt{RKNN\_QUERY\_MEM\_SIZE} across three 2-second steady-state trials; the board exposes no input-current sensor, so \rv power is omitted.

Across devices, \ours maintains a compact memory footprint alongside high throughput. On the \rv, \oursT, \oursS and \oursB require 24.9, 25.2 and 40.9\,MB of NPU DMA memory, compared with 53.6--80.5\,MB for UETrack-T/S and 49.8--187.4\,MB for HiT. On Orin Nano, \ours allocates only 11--18\,MB of TensorRT GPU memory (vs.\ 27--40\,MB for UETrack and 25--91\,MB for HiT) while sustaining 1203.9, 400.9 and 236.8\,FPS.

\begin{table*}[t]
  \centering\footnotesize
  \setlength{\tabcolsep}{4.8pt}
  \caption{\textbf{Accuracy, throughput, memory and power on edge devices.}
  AUC (\%) as well as \rv and Raspberry Pi~5 FPS follow \cref{tab:main}; Orin Nano FPS, \texttt{Mem} and \texttt{Power} are medians over three steady-state graph trials.
  \texttt{Mem} reports TensorRT GPU memory on Orin Nano, RKNN DMA memory on \rv, and process peak RSS on Raspberry Pi~5 (\rv has no on-board power sensor).
  Gray entries indicate that the RKNN FP16 graph failed the numerical check.}
  \label{tab:supp_runtime_resources}
  \begin{tabular}{@{}l rr rrr rr rrr@{}}
    \toprule
    & \multicolumn{2}{c}{AUC (\%)}
    & \multicolumn{3}{c}{Orin Nano (TensorRT)}
    & \multicolumn{2}{c}{\rv (RKNN)}
    & \multicolumn{3}{c}{Raspberry Pi~5 (ncnn)} \\
    \cmidrule(lr){2-3}\cmidrule(lr){4-6}\cmidrule(lr){7-8}\cmidrule(l){9-11}
    Tracker & LaSOT & VastTrack & FPS & Mem & Power & FPS & Mem & FPS & Mem & Power \\
    \midrule
    FEAR-XS & 50.1 & 29.6 & 561.7 & 7\,MB & 11.89\,W & 258.5 & 9.1\,MB & 91.6 & 51.4\,MB & 5.33\,W \\
    LightFC & 60.3 & 30.3 & 648.0 & 10\,MB & 12.44\,W & 135.4 & 18.0\,MB & 69.2 & 75.6\,MB & 5.43\,W \\
    LightFC-vit & 63.5 & 30.9 & 289.1 & 20\,MB & 16.55\,W & 21.5 & 55.7\,MB & 11.0 & 94.8\,MB & 4.96\,W \\
    HiT-Tiny & 55.1 & 25.7 & 272.0 & 50\,MB & 15.90\,W & 61.9 & 49.8\,MB & 27.8 & 116.5\,MB & 5.57\,W \\
    HiT-Small & 61.1 & 28.1 & 372.3 & 25\,MB & 11.47\,W & 43.9 & 60.0\,MB & 20.1 & 123.2\,MB & 5.11\,W \\
    HiT-Base & 65.3 & 30.0 & 216.7 & 91\,MB & 16.40\,W & 20.6 & 187.4\,MB & 7.1 & 258.6\,MB & 6.46\,W \\
    AsymTrack-T & 61.0 & 29.4 & 236.7 & 12\,MB & 10.73\,W & 52.4 & 26.7\,MB & 43.3 & 58.1\,MB & 5.04\,W \\
    AsymTrack-S & 62.6 & 30.2 & 219.1 & 12\,MB & 10.76\,W & 49.5 & 28.2\,MB & 36.6 & 74.6\,MB & 5.06\,W \\
    AsymTrack-B & 64.7 & 31.2 & 149.2 & 19\,MB & 12.55\,W & 19.6 & 45.1\,MB & 14.9 & 90.6\,MB & 5.17\,W \\
    FARTrack-Pico & 57.9 & 30.4 & 241.1 & 15\,MB & 11.96\,W & \textcolor{gray}{14.9} & \textcolor{gray}{38.0\,MB} & 18.0 & 64.7\,MB & 4.96\,W \\
    FARTrack-Nano & 59.5 & 34.0 & 105.2 & 48\,MB & 15.33\,W & \textcolor{gray}{10.4} & \textcolor{gray}{46.0\,MB} & 9.7 & 93.5\,MB & 5.10\,W \\
    FARTrack-Tiny & 61.7 & 35.1 & 71.5 & 67\,MB & 15.46\,W & \textcolor{gray}{7.5} & \textcolor{gray}{56.1\,MB} & 6.5 & 112.1\,MB & 4.86\,W \\
    UETrack-T & 63.4 & 33.2 & 454.6 & 27\,MB & 17.19\,W & 39.8 & 53.6\,MB & 14.3 & 98.5\,MB & 5.37\,W \\
    UETrack-S & 66.5 & 36.6 & 363.5 & 40\,MB & 17.74\,W & 28.9 & 80.5\,MB & 9.6 & 132.2\,MB & 5.30\,W \\
    MaST-Nano & 59.2 & 30.7 & 276.0 & 31\,MB & 16.46\,W & \textcolor{gray}{22.2} & \textcolor{gray}{35.3\,MB} & 22.6 & 82.6\,MB & 5.18\,W \\
    MaST-Tiny & 64.0 & 33.2 & 193.5 & 38\,MB & 14.70\,W & \textcolor{gray}{19.3} & \textcolor{gray}{44.1\,MB} & 16.6 & 89.4\,MB & 4.97\,W \\
    \midrule
    \rowcolor{oursrow}\oursT & 62.3 & 33.9 & 1203.9 & 11\,MB & 16.36\,W & 183.6 & 24.9\,MB & 96.3 & 67.1\,MB & 6.41\,W \\
    \rowcolor{oursrow}\oursS & 64.5 & 36.5 & 400.9 & 11\,MB & 12.82\,W & 85.3 & 25.2\,MB & 44.4 & 58.6\,MB & 6.02\,W \\
    \rowcolor{oursrow}\oursB & 65.6 & 37.0 & 236.8 & 18\,MB & 12.81\,W & 50.4 & 40.9\,MB & 27.0 & 70.6\,MB & 5.97\,W \\
    \bottomrule
  \end{tabular}
\end{table*}

\paragraph{HiT backbones with our interaction}
On LeViT-128S (\cref{tab:same_backbone}), caching the template shrinks the first two attention blocks from 320 to 256 tokens, which lowers fused-attention time from 5.94 to 4.15\,ms and the convolution and linear projections by 1.55\,ms.
The 11 attention operators and 5.71\,ms of layout and indexing overhead that remain in the LeViT backbone keep the per-frame graph at 13.49\,ms, against 5.45\,ms for \oursT.

\section{LaSOT attribute analysis}
\label{sec:supp_attributes}
\begin{figure}[tb]
  \centering
  \includegraphics[width=\linewidth]{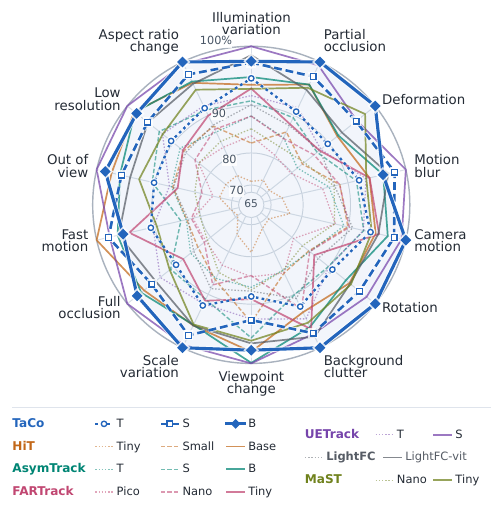}
  \caption{\textbf{LaSOT precision (P) by attribute.} Each spoke uses the test sequences tagged with that attribute. We measure center precision at 20 pixels and divide each tracker's score by the highest among the plotted trackers on that spoke. The radial scale spans 65--100\% of this maximum. Color identifies the tracker family and line style identifies variants within each family.}
  \label{fig:supp_lasot_attributes}
\end{figure}
\Cref{fig:supp_lasot_attributes} compares the seven multi-variant tracker families (18 variants) from \cref{tab:main} on the 14 LaSOT attributes.
\oursB has the highest precision on seven of them: partial occlusion, deformation, camera motion, rotation, background clutter, scale variation and aspect-ratio change.
Its margins over trackers of similar overall accuracy are largest on deformation, by 5.4, 5.6 and 4.8 points over AsymTrack-B, HiT-Base and LightFC-vit, and on rotation, by 4.3, 3.7 and 3.6 points.
These attributes change the outline of the target while most of its parts stay visible.
LightFC-vit also uses pixel-wise correlation but with kernels on a fixed grid, so the margin over it suggests that the box-aligned $4{\times}4$ target kernels help here, because each kernel can still match its own part when the shape changes. This is a hypothesis that the attribute split alone does not isolate.

UETrack-S leads on six of the remaining attributes and HiT-Base on fast motion.
\oursB trails UETrack-S by 0.9--2.6 points on out-of-view, full occlusion, low resolution, fast motion, viewpoint change, illumination variation and motion blur.
In these cases the target either departs from its first-frame appearance or covers only a few cells of the stride-16 grid, and \ours has neither a template update nor a finer correlation map with which to respond.
At the fast end, \oursT exceeds both LightFC and AsymTrack-T on all 14 attributes, by 0.7--4.8 and 1.3--5.6 points.

\section{Visualization details}
\label{sec:supp_vis}
The appendix maps of \cref{fig:supp_mechanism} and qualitative rollouts of \cref{fig:supp_qualitative} are measured on the adopted 300-epoch \oursT checkpoint ($128{\times}128$ template, $256{\times}256$ search, four $s_3$ modulation sites) across the 243 sequences of UAV123, UAVDT and DTB70, which are not used in training.

\paragraph{Kernel responses}
A response map is $\widetilde R_i$ after batch normalization and kernel weighting (\cref{eq:kcond,eq:supp_bn_kcond}) on the stride-16 search grid.
Each of the 32 maps is standardized over its interior cells (excluding the outermost ring). The maps of the 16 target kernels, or of the 16 context kernels, are then averaged, and the signed mean is drawn after bilinear upsampling using a shared symmetric scale for all target and context tiles in each figure.
Red and blue denote positive and negative standardized responses, respectively, with zero corresponding to each kernel's spatial mean.
The outermost cells are not drawn, because the corner cell carries a padding artifact of the backbone that would set the scale.
The centroid of a kernel row is the centroid of the positive part of the mean map of its four kernels, over the cells whose centre lies in the predicted box enlarged by half a cell.

\paragraph{Mean modulation operands}
The mean scale and shift of every modulation site are taken over the 243 sequences.
To isolate per-frame selectivity without accumulating closed-loop drift, each search crop selected by the unmodified tracker is evaluated a second time with the mean operands, and both predictions are compared with the ground truth on the 162,444 annotated frames.
Even in this open-loop single-frame re-evaluation, where errors cannot propagate across frames, mean IoU drops from 0.679 with the target's operands to 0.677 with the mean operands.
We count a frame as losing the target when one prediction has IoU below 0.2 while the other is above 0.5: this happens with the mean operands on 175 frames versus 70 frames with the target operands.
The last two rows of \cref{fig:supp_mechanism}b show examples of the second kind.

\paragraph{Qualitative comparison}
\Cref{fig:supp_qualitative} extends \cref{fig:qualitative} to four frames per sequence, at 15, 40, 65 and 90\% of its length.
In the two failures, \oursT loses a person after 141 frames while the competing trackers keep it, and loses a car after about 800 frames that AsymTrack-T keeps.

\begin{figure*}[p]
  \centering
  \includegraphics[width=\linewidth]{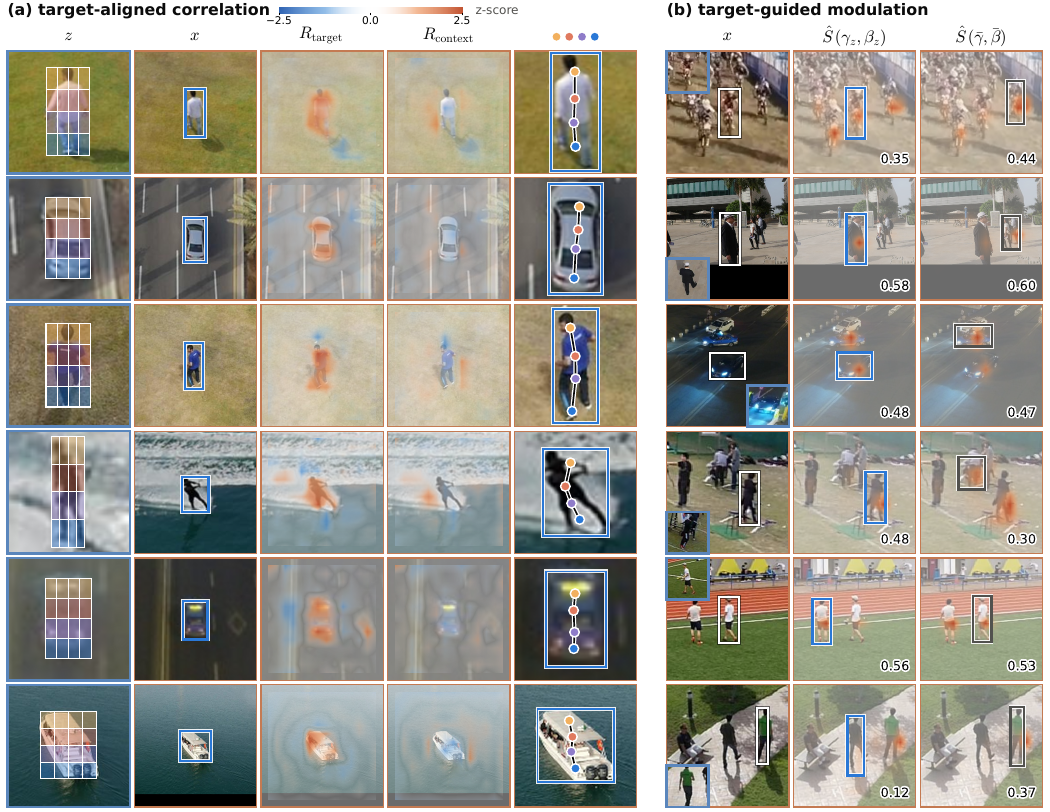}
  \caption{\textbf{Further measured maps of \oursT}, in the layout of \cref{fig:mechanism}.
  In the last two rows of (b), the prediction with the target's own operands is the one that loses the target.}
  \label{fig:supp_mechanism}
\end{figure*}

\begin{figure*}[p]
  \centering
  \includegraphics[width=\linewidth]{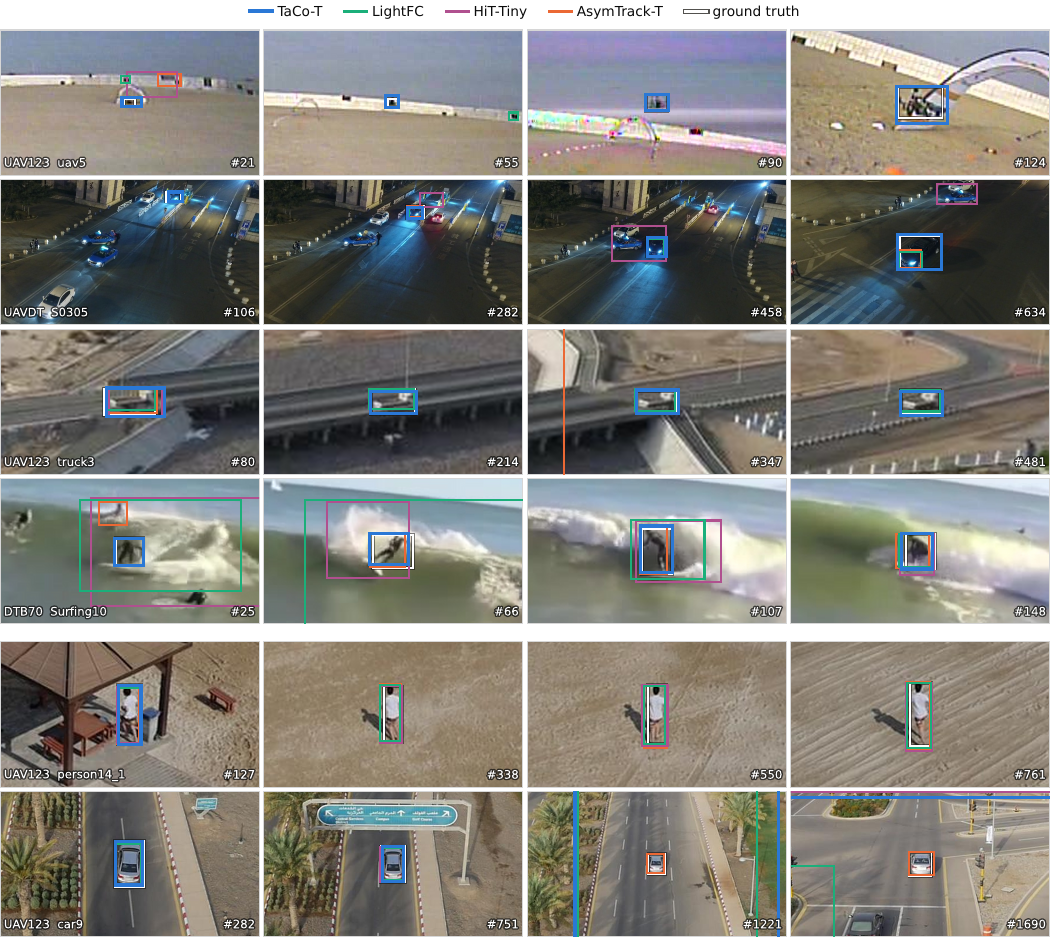}
  \caption{\textbf{Qualitative comparison on UAV123, UAVDT and DTB70}, with frames at 15, 40, 65 and 90\% of each sequence.
  \oursT keeps the target in the top four rows and loses it in the bottom two.}
  \label{fig:supp_qualitative}
\end{figure*}

\clearpage
{
  \small
  \putbib[main]
}
\end{suppbib}

\end{document}